\documentclass{article}

\usepackage{microtype}
\usepackage{graphicx}
\usepackage{subfigure}
\usepackage{subcaption}
\usepackage{booktabs} 

\usepackage{hyperref}
\usepackage{multirow}
\usepackage{tabularx}

\usepackage[accepted]{icml2025}

\usepackage{amsmath}
\usepackage{amssymb}
\usepackage{mathtools}
\usepackage{amsthm}

\usepackage[capitalize,noabbrev]{cleveref}

\theoremstyle{plain}

\theoremstyle{definition}

\theoremstyle{remark}

\usepackage[textsize=tiny]{todonotes}

\icmltitlerunning{Enhancing RAG with Active Learning on Conversation Records}

\begin{document}

\twocolumn[
\icmltitle{Enhancing RAG with Active Learning on Conversation Records:\\ Reject Incapables and Answer Capables}



\icmlsetsymbol{equal}{*}

\begin{icmlauthorlist}
\icmlauthor{Xuzhao Geng}{hust}
\icmlauthor{Haozhao Wang}{hust}
\icmlauthor{Jun Wang}{wudao}
\icmlauthor{Wei Liu}{hust}
\icmlauthor{Ruixuan Li}{hust}
\end{icmlauthorlist}

\icmlaffiliation{hust}{School of Computer Science and Technology, Huazhong University of Science and Technology, Wuhan, China}
\icmlaffiliation{wudao}{iWudao Tech, Hangzhou, China}

\icmlcorrespondingauthor{Xuzhao Geng}{geng\_xz@hust.edu.cn}
\icmlcorrespondingauthor{Haozhao Wang}{hz\_wang@hust.edu.cn}

\icmlkeywords{Machine Learning, ICML}

\vskip 0.3in
]



\printAffiliationsAndNotice{}  

\begin{abstract}
Retrieval-augmented generation (RAG) is a key technique for leveraging external knowledge and reducing hallucinations in large language models (LLMs). However, RAG still struggles to fully prevent hallucinated responses. To address this, it is essential to identify samples prone to hallucination or guide LLMs toward correct responses, which experts then annotate to develop high-quality datasets for refining LLMs. However, the growing scarcity of such datasets makes their creation challenging. This paper proposes using the vast amount of conversations from widespread LLM usage to build these datasets, training LLMs to avoid hallucination-prone questions while accurately responding to manageable ones. Given the impracticality of expert-annotating all conversation records, the paper introduces AL4RAG, which uses active learning to select the most suitable conversation samples for annotation, optimizing performance within an annotation budget. Additionally, recognizing that traditional active learning methods are not fully compatible with RAG due to unsuitable distance metrics, we develop a novel sample distance measurement for RAG active learning. Extensive experiments show that our method consistently outperforms baselines across multiple metrics.
\end{abstract}

\begin{figure*}[thbp!]
\vskip 0.1in
    \centering
    \begin{tabular}{@{\extracolsep{\fill}}c@{}c@{\extracolsep{\fill}}}
            \includegraphics[width=0.53\linewidth]{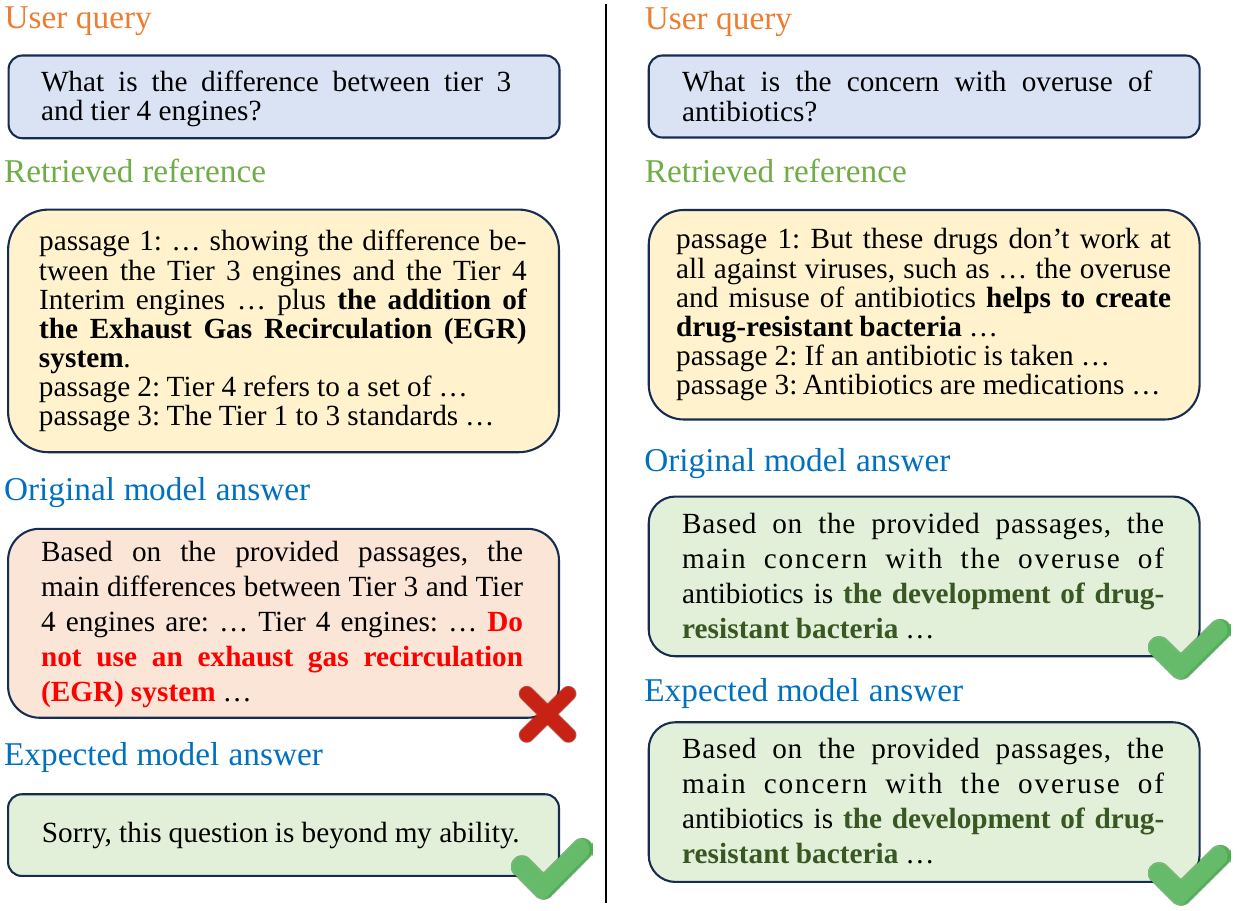} &
            \hspace{0.1cm}
            \includegraphics[width=0.46\linewidth]{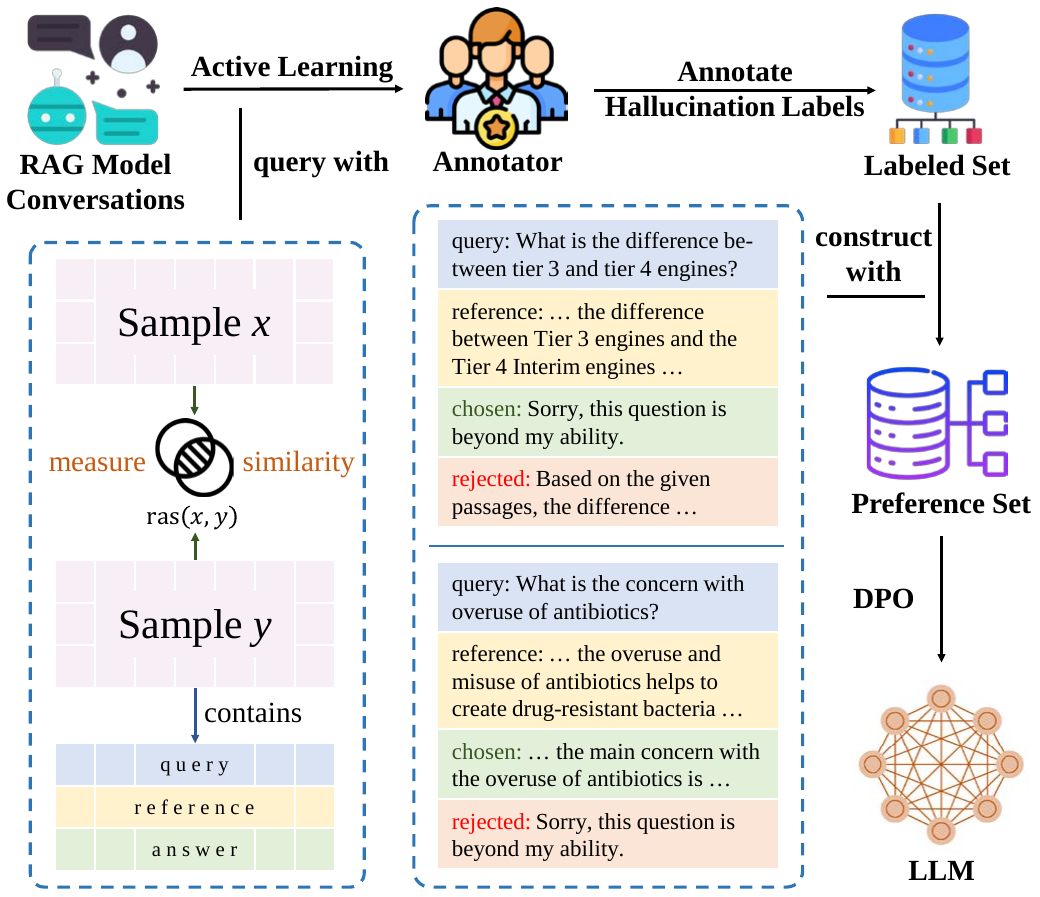}\\
            (a) An example related to our tasks & (b) Overall framework\\
    \end{tabular}
    \caption{(a) An example regarding our tasks. In the scenario on the left, the original model provides an incorrect response to the user; in this case, we expect the model to decline answering the question, whereas in the scenario on the right, where the original model delivers a correct response, we aim for it to generate accurate responses more consistently. (b) Overall framework of our approach. The example of preference set construction is the same as (a).}
    \label{fig:intro}
\vskip -0.1in
\end{figure*}

\section{Introduction}
In recent years, large language models (LLMs) have demonstrated remarkable performance in diverse natural language processing (NLP) tasks, such as text classification \cite{abburi2023generative}, summarization \cite{jin2024comprehensive}, and question answering \cite{zhuang2023toolqa}. However, they frequently encounter the issue of hallucinations \cite{huang2023survey}, which undermines the reliability of their responses. Retrieval-Augmented Generation (RAG) \cite{lewis2020retrieval}, integrated into leading models like GPT-4 \cite{achiam2023gpt}, Gemini-1.5 \cite{team2024gemini}, and Claude-3.5 \cite{anthropic2024claude}, aims to tackle this problem. RAG combines a retriever to fetch relevant documents and a generator to formulate answers, enhancing the model's reliability by leveraging external knowledge. Nevertheless, RAG cannot completely eliminate hallucinations \cite{chen2024benchmarking,wood2024100}, posing a persistent challenge in ensuring the quality of model-generated content.

The challenge of addressing hallucinations highlights the need for systems that can effectively identify and manage situations prone to hallucination. Our research aims to train models to reject hallucination-prone queries while ensuring stable and accurate responses to queries within their capability, as illustrated in Figure \ref{fig:intro}(a). Achieving this goal requires high-quality model conversation datasets specifically designed to teach the model when to refuse to answer, while also including queries that the model can answer correctly in order to maintain its capabilities. However, relevant datasets are exceedingly scarce, and the creation of such datasets is highly challenging due to the need for extensive manual annotation, while the majority of existing model conversation data remains unlabeled. To address this, we propose leveraging active learning (AL) \cite{settles1995active} to screen model conversation records. AL selects the most informative data from large pools of unlabeled data, enabling the creation of a small but high-quality human-annotated dataset through focused manual annotation.

Unfortunately, existing AL methods perform poorly in the RAG scenario. These methods can be categorized into uncertainty-based and diversity-based approaches. Traditional uncertainty-based methods demonstrate instability when applied to RAG-related tasks \cite{tsvigun2022active, snijders2023investigating}, while existing diversity-based methods \cite{maekawa2022low,xie2023active} fail to consider the unique three-segment structure of RAG conversation records (e.g., a query, retrieved documents, and a model-generated response), resulting in insufficient diversity in the collected data. With the advent of LLMs, some researchers have employed large models to measure sample uncertainty \cite{li2024active}, but this approach requires large models to evaluate the uncertainty of all unlabeled samples sequentially, leading to excessive time and computational costs when applied in practical RAG scenarios. Other researchers have used LLMs to replace human annotators for labeling samples \cite{zhang2023llmaaa,xiao2023freeal}, but the labeling accuracy of LLMs is far inferior to that of human annotators, resulting in lower-quality datasets. Therefore, developing AL methods tailored specifically for RAG systems becomes crucial.

To address this gap, this paper proposes \textbf{AL4RAG}, a novel AL strategy specifically designed for RAG models, as illustrated in Figure \ref{fig:intro}(b). We improve upon existing diversity-based methods \cite{tsvigun2022active,margatina2023active} by independently considering the various fields of RAG data, which effectively accommodates the unique nature of RAG data. This approach selectively identifies the most diverse samples from unlabeled model conversation records. By annotating these strategically selected samples, we can construct a high-quality human-annotated preference dataset. However, to construct a preference dataset, it is essential to introduce a refusal option. Yet, in existing RAG model conversation datasets, such as RAGTruth \cite{niu2023ragtruth}, each question corresponds to only a single answer, allowing for the identification of hallucinations but lacking the option to refuse to answer. To address this challenge, we ask human annotators to assign hallucination labels to the selected model conversations. Each sample is then labeled based on the presence of hallucinations, indicating whether the model should prefer to answer or refuse to respond. This approach enables us to build a dedicated preference dataset. Using this dataset for model optimization can significantly improve the performance of RAG models, thereby achieving our intended objectives. Our main contributions are:

\begin{itemize}
\item To the best of our knowledge, we are the first to propose an AL framework for RAG, presenting an effective selection strategy in response to the unique data pattern of RAG data. 
\item We proposed retrieval-augmented similarity (\textbf{\textit{ras}}) for measuring the similarity between samples within the unique patterns of RAG data, enabling more accurate measurement of sample distances.
\item We expanded the RAGTruth dataset and created the first human preference dataset tailored to the RAG scenario for handling both hallucination-prone queries and answerable ones.
\item Extensive AL-driven model optimization was conducted on the constructed dataset, with results demonstrating the effectiveness of our approach.
\end{itemize}

\section{Related Work}

\subsection{Active Learning}

Active learning (AL) is a widely adopted technique for optimizing the trade-off between annotation costs and model performance by selecting the most informative samples from large unlabeled datasets. Central to AL are three components: a labeling oracle, an unlabeled data pool, and a query strategy. Common strategies include uncertainty-based methods, which prioritize difficult samples based on prediction uncertainty \cite{beluch2018power,liu2021influence,schroder2021revisiting,maekawa2022low,rouzegar2024enhancing}, and diversity-based methods, which focus on selecting diverse samples to enrich datasets \cite{hasan2018context,sinha2019variational,agarwal2020contextual,maekawa2022low,xie2023active}. 

Active learning (AL) has been effectively applied across various NLP tasks, including text classification \cite{yan2020active,schroder2021revisiting}, text summarization \cite{gidiotis2022should,tsvigun2022active}, and question answering \cite{karamcheti2021mind,padmakumar2021dialog}, achieving significant cost reductions and performance improvements. These approaches have demonstrated strong potential in optimizing model training efficiency and enhancing overall system performance. Despite their successes, existing methods often neglect the influence of inherent sample properties on diversity. Addressing this gap, our work introduces a novel approach for evaluating sample diversity in the RAG context by comparing similarities across different data fields.

\subsection{Active Learning Meets LLMs}

As large language models (LLMs) continue to advance, their integration with AL has become a focal point for addressing high annotation costs \cite{tan2024large} and challenges in effective knowledge utilization \cite{xu2024activerag}. Currently, the integration of AL with LLMs primarily involves three approaches: employing traditional active learning methods to select samples for the downstream processes of LLMs (e.g., fine-tuning, in-context learning, evaluation) \cite{xie2023active,margatina2023active,bayer2024activellm}, utilizing LLMs to assess sample quality (e.g., uncertainty estimation) \cite{li2024active}, and leveraging LLMs to replace human annotators \cite{xiao2023freeal,kholodna2024llms}. For instance, Margatina et al. \yrcite{margatina2023active} demonstrated the effectiveness of similarity sampling for classification, framing in-context learning’s example selection as a single-round AL task. Li et al. \yrcite{li2024active} proposed LDCAL for text summarization, while Rouzegar and Makrehchi \yrcite{rouzegar2024enhancing} balanced cost and accuracy in text classification. Other studies addressed noisy data filtering \cite{taneja2024can} and explored the use of LLMs as annotators \cite{zhang2023llmaaa}, highlighting both strengths and limitations.

Our research integrates AL into the RAG framework, leveraging its capabilities to address the unique challenges of fine-tuning LLMs. Specifically, we focus on selecting high-impact samples that enhance model performance while considering diversity within the RAG setting. To the best of our knowledge, this is the first study to explore AL-driven optimization for LLMs in the RAG context.

\section{Problem Definition \& Preliminaries}

\subsection{Problem Definition}

To enhance a model's performance in the RAG setting, we utilize preference optimization using its historical conversations. We aim to enable the model to reject queries likely to cause hallucinations and to improve its stability in answering queries within its capability. Each unlabeled sample comprises a user query, a reference document, and a model-generated response. We identify and label informative samples by determining whether the responses exhibit hallucinations and use these labeled samples for preference optimization to refine the model's decision-making.

Inspired by prior works \cite{tsvigun2022active,margatina2023active}, we adopt a diversity-based approach, evaluating diversity by computing distances among samples to prioritize varied and informative data.

\subsection{Diversity-based Active Learning}

Given an unlabeled dataset \( U = \{x_1, x_2, \dots, x_n\} \), feature extraction algorithms (e.g., TF-IDF) transform each sample \( x_i \) into a feature vector \( \mathbf{x_i} \). Initially, \( k \) samples are randomly selected to form the initial selected set \( S = \{s_1, s_2, \dots, s_k\} \), and these samples are removed from \( U \).  

In subsequent rounds, the average distance \( \mathcal{D} \) between each sample \( x_i \) in \( U \) and all samples in \( S \) is measured as:
\begin{equation}
  \label{eq:distance}
    \mathcal{D}(x_i)=\frac{1}{|S|}\sum_{j = 1}^{|S|}(1-\frac{\mathbf{x_i} \cdot \mathbf{s_j}}{\|\mathbf{x_i}\| \|\mathbf{s_j}\|})
\end{equation}
The top-\( k \) samples ranked by \( \mathcal{D} \) are added to \( S \) and subsequently removed from \( U \). This process repeats until the annotation budget is reached.  

Unlike methods that label samples during each selection round, we label all samples in \( S \) collectively at the end, reducing annotators' waiting time and improving efficiency.

\subsection{RAG Framework}

In the domain of NLP, RAG is a crucial methodology. Suppose a user inputs a natural language query \( q \), there is a pre-established knowledge repository housing a collection of text chunks \( r_i \). The system computes a set of relevance scores \( f(q, r_i) \) to screen relevant text chunks from this repository against $q$. Next, based on a predefined threshold \( \tau \) or a predefined number \( k \), the text chunks that meet the condition $f(q, r_i)>\tau$ or the top-$k$ relevant chunks are retrieved and ranked, generating a set \( R = \{r_1, r_2, \dots, r_k\} \). These retrieved chunks are combined with \( q \) into a prompt \( P = [q; R] \), which is the input to an LLM \( \mathcal{M} \) to generate an answer \( a = \mathcal{M}(P) \). By leveraging the retrieved text, the generated answer attains enhanced contextual accuracy.

Evidently, in contrast to other scenarios, within the RAG framework, each conversation of the model encompasses multiple attributes, specifically the user-posed query, the retrieved references, and the answer generated by the model. In this setting, the measurement of distances between samples becomes more complicated. The direct application of AL can lead to inaccurate measurement of distances among samples. Hence, it is imperative to develop a method for measuring sample distances tailored to the RAG scenario.

\section{AL4RAG}

In order to train models to handle both hallucination-prone queries and valid ones with limited annotation budget, we need to select a subset of the most informative samples from the model's historical conversation records. Next, we will annotate these samples and construct a preference dataset. Finally, we use these annotated samples for DPO training. The specific process is as follows:

\subsection{Active Learning Process}

We employ AL to select informative samples from the RAG model's historical conversation records. Specifically, the process entails the following steps:

(1) Random selection of initial samples for labeling. Initially, a small subset of samples is selected randomly from the entire dataset as the selected set.

(2) Measurement of similarity between remaining and selected samples. The second step involves measuring the similarity between unselected samples and the selected ones. In this step, we use the user input (e.g., user query, question) $q$ as the unit for measuring sample similarity.
\begin{equation}
  \label{eq:inputsimilarity}
    \operatorname{sim}(x, y) = \frac{\mathbf{x_q} \cdot \mathbf{y_q}}{\|\mathbf{x_q}\| \|\mathbf{y_q}\|}
\end{equation}
where $\mathbf{x_q}$, $\mathbf{y_q}$ refer to queries of sample $x$ and $y$.

(3) Measurement of similarity among remaining samples. The same method is used to measure the similarity among the remaining samples in the unselected set.

(4) Scoring of remaining samples. An IDDS score \cite{tsvigun2022active} is assigned to each of the remaining samples based on the equation as follows: 
\begin{equation}
\operatorname{IDDS}(x) = 
\lambda \frac{\sum_{j=1}^{|U|} \operatorname{sim}(\mathbf{x}, \mathbf{x_j})}{|U|} 
- (1 - \lambda) \frac{\sum_{i=1}^{|S|} \operatorname{sim}(\mathbf{x}, \mathbf{x_i})}{|S|}
\end{equation}
where $\lambda$ is a hyper-parameter, $U$ refers to the unselected set and $S$ refers to the selected set.

(5) Expansion of the selected set: Move the top \%k samples, based on their scores, to the selected set. Then, proceed back to step (2).

Our goal is to select samples with low similarity to already selected ones and high similarity to unselected ones, thereby excluding outliers. This process iterates until the target sample count is achieved.

\subsection{Preference Dataset Construction}\label{step2}

After obtaining an informative subset through AL, we need to annotate these samples to construct a preference dataset. In general, the construction of a preference dataset involves the annotator choosing the better one among two candidate answers $(a_1, a_2)$ to an input prompt $p$. Then, the better one is set as $a_w$, and the other one is set as $a_l$. However, in practical applications of the RAG model, users typically request only a single response from the model for a given query. As a result, we are unable to obtain a pair of answers to a query. For our task, we propose a novel method for dataset construction. Specifically, we ask annotators to assess whether the model's response in each RAG conversation contains hallucinations and then generate a label $h$, where $h = 0$ indicates that the response is free of hallucinations, and $h = 1$ indicates that the response contains hallucinations. Next, we examine the \( h \) label of each sample. For samples where \( h = 0 \), we designate the model's original response as \( a_w \) and an explicit rejection response as \( a_l \). Conversely, for samples where \( h = 1 \), we assign the explicit rejection response as \( a_w \) and the model's original response as \( a_l \). By modifying the feedback strategy, we successfully construct a preference dataset tailored for scenarios involving only a single model response.

\subsection{Fine-tuning Process}\label{step3}

Inspired by previous work \cite{khaki2024rs}, we adopt DPO \cite{rafailov2024direct} to achieve the aforementioned objectives. DPO fine-tunes LLMs to align their outputs with human preferences, simplifying optimization by eliminating the need for complex reward function estimation \cite{ouyang2022training,bai2022training}.  

With the constructed preference dataset, we define two policies: $\pi_{\theta}$, the model being optimized, and $\pi_{o}$, the original model used as a baseline. The optimization phase involves minimizing a loss function based on human preferences:
\begin{equation}
\begin{aligned}
    \mathcal{L}_\text{DPO}(\theta) \!=\! 
    -\!\log \sigma\beta\left( \log \frac{\pi_{\theta}(a_w|p)}{\pi_{o}(a_w|p)} - \log \frac{\pi_{\theta}(a_l|p)}{\pi_{o}(a_l|p)}\right)
\end{aligned}
\end{equation}
where $\sigma$ is the sigmoid function, and $\beta$ adjusts alignment speed with human preferences.

By fine-tuning the model with DPO, we can achieve the previously outlined optimization objectives, thereby mitigating model trust issues caused by hallucinations.

\section{AL4RAG\textsubscript{ras}}
\subsection{Observation}
\textit{Conventional sample similarity measurement based on user input can lead to inaccurate judgment of sample distances.}

Classical Active Learning (AL) has achieved significant success in both Computer Vision (CV) \cite{budd2021survey,tuia2021survey,wang2023comprehensive} and Natural Language Processing (NLP) \cite{zhang2022survey}, with numerous methods leveraging sample diversity to enhance learning efficiency \cite{shi2021diversity,kim2021task,li2022batch,jin2022one}. In CV tasks, diversity-based queries often involve directly computing distances between input images, enabling efficient selection of informative samples. Similarly, in NLP tasks such as text classification \cite{tan2021diversity}, text summarization \cite{tsvigun2022active}, semantic parsing \cite{li2023best}, and information extraction \cite{duan2024research}, distance measurements can also be applied directly, as samples typically consist of only a single attribute—the input text itself. This simplicity allows for straightforward similarity measurements, making traditional AL strategies effective in these domains.

\begin{figure}[t]
    \centering
    \includegraphics[width=\columnwidth]{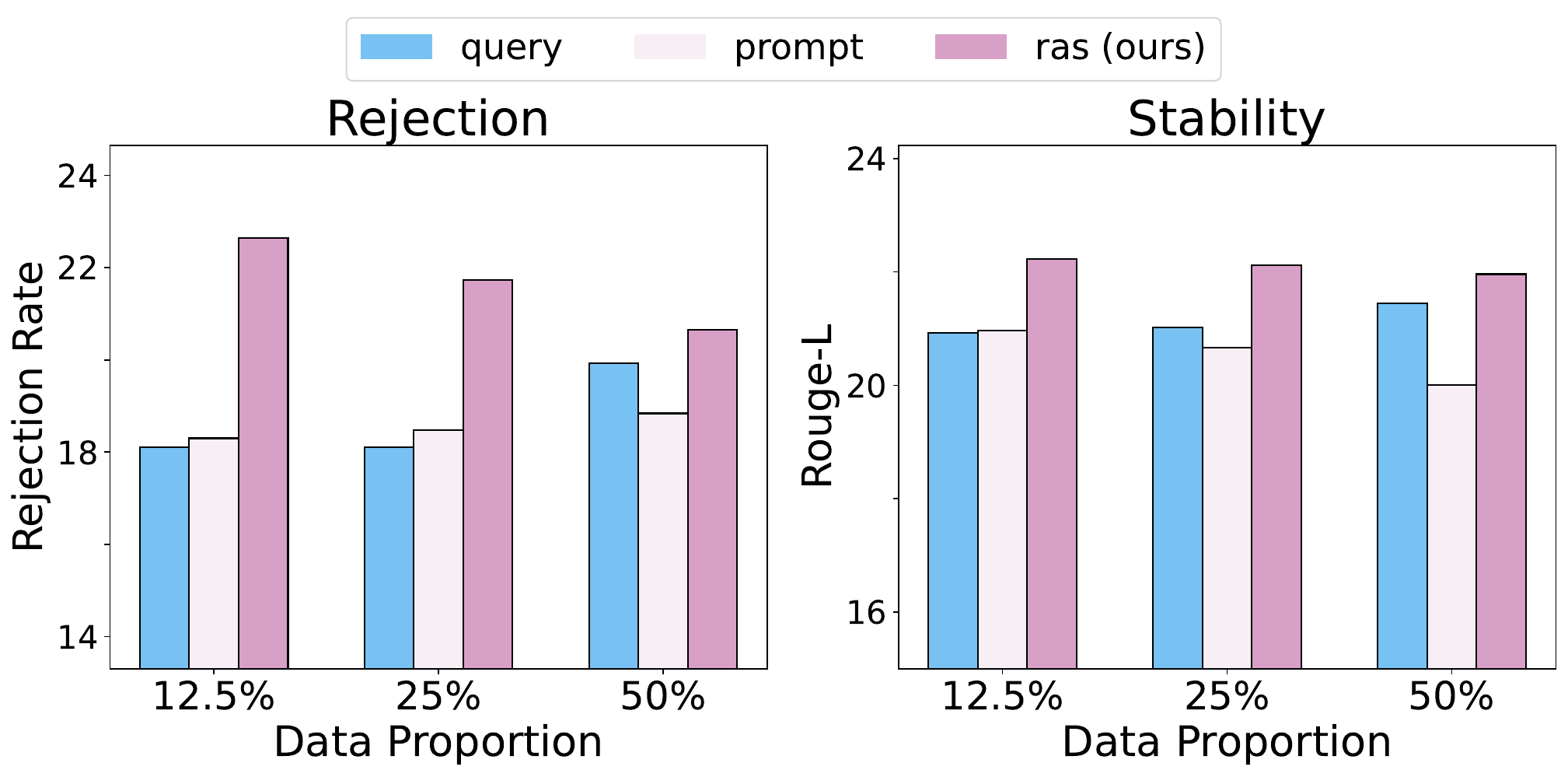}
    \caption{Part of the experimental results of query similarity, prompt similarity and \textit{ras}, which shows the impact of different similarity measurement on model performance. The left graph shows the performance of handling hallucination-prone queries, and the graph on the right shows the performance of handling model-answerable queries.}
    \label{fig:motivation}
\end{figure}

In contrast, the RAG scenario introduces conversation records characterized by three distinct attributes: the user query, the reference documents retrieved by the retriever, and the LLM-generated response. Measuring sample similarity solely based on user input text can lead to inaccuracies, as RAG hallucination is usually caused by the misunderstanding of retrieved references \cite{chen2024benchmarking}. However, if we consider combining multiple attributes (e.g., queries and referneces) or relying on prompt-based similarity (i.e., combining queries, references, and template text), it can also result in inaccuracies due to variations in the lengths of these attributes. For instance, the retriever may retrieve similar reference documents for two different user queries. When such measurement method is applied, the similarity between these two samples is likely to be disproportionately high, as the lengths of the reference documents and the template text dominate, overshadowing the much shorter query. This makes the distance between the two samples artificially small, rendering them nearly indistinguishable, even though the query itself plays a critical role in determining the model's performance. To address this, our approach evaluates each attribute—query, reference, and response—independently, enabling more accurate similarity measurements and clearer differentiation between samples. This ensures that the unique influence of the query is appropriately captured. Figure \ref{fig:motivation} demonstrates that our method consistently outperforms input-based similarity and prompt-based similarity across varying data ratios and tasks. Further detailed results are provided in Section \ref{sec:comparison}. This highlights the robustness and adaptability of our approach to real-world RAG applications.

\subsection{AL4RAG with Retrieval-Augmented Similarity}
Similar to the method mentioned previously, we carry out the same series of steps to screen samples. However, in step (2) and (3), to address the observed issues, we propose \textbf{retrieval-augmented similarity} (\textit{ras}), a novel similarity measurement method specifically designed for the RAG scenario. Here, each sample consists of three attributes: a user-input query $q$, a reference $r$ combining multiple retrieved documents, and a model-generated answer $a$. The query and reference are then combined into a prompt $p$, serving as the model's input, while the answer represents the output.  

Unlike typical AL algorithms, which often treat the input as a single entity \cite{margatina2023active, taneja2024can}, \textit{ras} considers each attribute independently. We measure the similarity for $q$ and $r$ separately, and obtain the average of these values. After that, we choose the minimum between the obtained value and the similarity of $p$ as the final similarity. This nuanced method ensures a more accurate and context-aware assessment of sample similarity. Let $x$ and $y$ be two different samples, $ras(x,y)$ refers to the retrieval-augmented similarity between them. It is measured as follows:
\begin{equation}
  \label{eq:similarity}
    \operatorname{ras}(x, y) = \operatorname{min}(\frac{\mathbf{x_p} \cdot \mathbf{y_p}}{\|\mathbf{x_p}\| \|\mathbf{y_p}\|}, \frac{1}{2} (\frac{\mathbf{x_q} \cdot \mathbf{y_q}}{\|\mathbf{x_q}\| \|\mathbf{y_q}\|} + \frac{\mathbf{x_r} \cdot \mathbf{y_r}}{\|\mathbf{x_r}\| \|\mathbf{y_r}\|}))
\end{equation}
where $\mathbf{x_p}$, $\mathbf{y_p}$, $\mathbf{x_q}$, $\mathbf{y_q}$ and $\mathbf{x_r}$, $\mathbf{y_r}$ refer to $p$, $q$ and $r$ of $x$ and $y$. In the main experiments of this paper, we utilized TF-IDF vectorization. In the subsequent experiments, we compared the impacts of different vectorization methods on the final performance of the model.

After the AL process, we construct the preference dataset and fine-tune the model in the same way as described in Section \ref{step2} and \ref{step3}.

\section{Experiments}

\subsection{Experimental Settings}

\textbf{Tasks \& Dataset}~~
We evaluated our method on Question Answering, Summary, and Data-to-text writing tasks using the RAGTruth dataset \cite{niu2023ragtruth}, a high-quality, manually annotated dataset for RAG hallucination. The dataset includes queries, references, model responses, and hallucination labels. To prepare the preference data, we identified the hallucination labels of responses generated by the Llama-2-7B-chat model \cite{touvron2023llama} and assigned preferred and non-preferred answers accordingly. The resulting dataset, containing approximately 3,000 entries, was split into training and test sets, with each sample comprising a query, a reference, a chosen answer, and a rejected answer. AL strategies are applied on the training set only.

\textbf{Model}~~
The standard DPO process relies on an SFT model, but none exists for our task. Thus, we fine-tuned the llama-2-7B-chat model on the training set. This enabled the model to better assess query complexity and appropriately handle scenarios requiring rejection responses.

\textbf{Baselines}~~
To verify the effectiveness of our method, we selected several baselines for comparison, including:

\begin{itemize}
\item \textbf{Random}: The samples for annotation are randomly selected all at once.
\item \textbf{Entropy} \cite{wang2014new}: An AL strategy that selects samples with the highest prediction uncertainty, measured by maximum entropy, to improve model performance.
\item \textbf{Coreset} \cite{sener2017active}: Select a small, representative subset of data that approximates the entire dataset’s distribution for efficient learning.
\item \textbf{BLEUVar} \cite{xiao2020wat}: Use BLEU variance to prioritize samples with high uncertainty by treating documents as points in a high-dimensional space.
\item \textbf{Naive IDDS} \cite{tsvigun2022active}: Select samples dissimilar to labeled ones but similar to unlabeled ones, based on document embeddings.
\item \textbf{IDDS}\textsubscript{prompt}: Use prompt similarity, incorporating queries, references, and template text, as the sample similarity measure, with IDDS as the query strategy.
\begin{table}[h]
\centering
\caption{Overall rejection performance of different algorithms under different data proportions. The best results are \textbf{bolded}, and the second-best results are \underline{underlined}.}
\vskip 0.1in
\begin{tabular}{lccccccccccc}
\toprule
\textbf{Algorithms} & \textbf{12.5\%} & \textbf{25\%} & \textbf{50\%} \\
\midrule
Random & 17.39 & 16.85 & 19.02 \\
Entropy & 18.48 & \underline{19.74} & \underline{19.93} \\
Coreset & 18.30 & 17.03 & \underline{19.93} \\
BLEUVar & 18.11 & 18.84 & \underline{19.93} \\
IDDS & 18.66 & 19.02 & 19.57 \\
IDDS\textsubscript{prompt} & 18.30 & 18.48 & 18.84 \\
IDDS\textsubscript{q,r combined} & \underline{18.84} & 19.57 & 17.03 \\
\midrule
AL4RAG & 18.11 & 18.11 & \underline{19.93} \\
AL4RAG\textsubscript{ras} & \textbf{22.65} & \textbf{21.74} & \textbf{20.65} \\
\bottomrule
\end{tabular}
\label{tab:rej_performance}
\end{table}
\item \textbf{IDDS}\textsubscript{q,r combined}: Similar to the above method, but only concatenate the query and the reference, and use the similarity of this part as the sample similarity.
\end{itemize}

\textbf{Evaluation Metrics}~~
For the performance of refusing to answer queries, we use \textbf{Rejection Rate (RR)} for evaluation. For the performance of correctly answering queries, we measure it by the text similarity between the model's responses and the correct reference answers, including:

\begin{itemize}
\item \textbf{ROUGE Family:} Widely used metrics for measuring text similarity. \textbf{ROUGE-L} gauges generation quality by measuring similarity between generated and reference summaries. \textbf{ROUGE-1} counts overlapping words or n-grams for assessment. \textbf{ROUGE-2} determines similarity via co-occurring word bigrams, offering another way to quantify summary matching. 
\item \textbf{BERTScore:} A fine-grained, context-aware NLP evaluation metric that uses BERT to measure the cosine similarity between token embeddings of candidate and reference texts.
\end{itemize}

\textbf{Implementation Details}~~
For the AL process, we set the number of iteration rounds for both AL4RAG and IDDS to five. In the fine-tuning process, we employ LoRA \cite{hu2021lora} with a learning rate of \(1 \times 10^{-5}\) and train for one epoch. During the generation phase, we set the temperature of the LLM to 0.7 and report the mean performance across five independent generation runs.

\subsection{Main Results} \label{sec:comparison}

\subsubsection{Rejection Results}

\begin{table*}[h]
\centering
\caption{Overall stability performance of different algorithms under different data proportions. The best results are \textbf{bolded}, and the second-best results are \underline{underlined}.}
\vskip 0.1in
\normalsize
\resizebox{\textwidth}{!}{
\begin{tabular}{lcccccccccccc}
\toprule
\multirow{2}{*}{\textbf{Algorithms}} & \multicolumn{4}{c}{\textbf{12.5\%}} & \multicolumn{4}{c}{\textbf{25\%}} & \multicolumn{4}{c}{\textbf{50\%}} \\
\cmidrule(lr){2-5} \cmidrule(lr){6-9} \cmidrule(lr){10-13}
 & \scriptsize\textbf{Rouge-L} & \scriptsize\textbf{Rouge-1} & \scriptsize\textbf{Rouge-2} & \scriptsize\textbf{BERTScore} & \scriptsize\textbf{Rouge-L} & \scriptsize\textbf{Rouge-1} & \scriptsize\textbf{Rouge-2} & \scriptsize\textbf{BERTScore} & \scriptsize\textbf{Rouge-L} & \scriptsize\textbf{Rouge-1} & \scriptsize\textbf{Rouge-2} & \scriptsize\textbf{BERTScore} \\
\midrule
Random & 20.68 & 29.02 & 14.41 & 28.34 & 20.07 & 28.38 & 13.86 & 27.66 & 20.52 & 29.14 & 14.50 & 28.36 \\
Entropy & 20.60 & 29.28 & 14.40 & 29.09 & 21.59 & 30.15 & 15.25 & \underline{30.00} & 20.37 & 29.08 & 14.08 & 28.10 \\
Coreset & 20.30 & 28.92 & 14.18 & 28.49 & 20.78 & 29.01 & 14.50 & 28.67 & 20.69 & 29.27 & 14.63 & 28.69 \\
BLEUVar & 19.75 & 28.48 & 13.65 & 27.38 & \underline{21.62} & \underline{30.46} & \underline{15.33} & 29.89 & 20.46 & 28.99 & 14.13 & 28.21 \\
Naive IDDS & 20.49 & 29.07 & 14.47 & 28.75 & 20.65 & 29.06 & 14.24 & 29.05 & 19.80 & 27.83 & 13.85 & 27.67 \\
IDDS\textsubscript{prompt} & \underline{20.96} & 29.69 & \underline{14.88} & 28.90 & 20.67 & 29.15 & 14.44 & 28.64 & 20.00 & 28.79 & 13.95 & 28.02 \\
IDDS\textsubscript{q,r combined} & 20.70 & 29.18 & 14.38 & 28.47 & 19.92 & 28.10 & 13.66 & 27.52 & 20.69 & 28.95 & 14.35 & 28.18\\
\midrule
AL4RAG & 20.93 & \underline{29.74} & 14.57 & \underline{29.60} & 21.02 & 29.57 & 14.61 & 29.37 & \underline{21.44} & \underline{30.64} & \underline{14.81} & \underline{29.64} \\
AL4RAG\textsubscript{ras} & \textbf{22.23} & \textbf{31.17} & \textbf{15.63} & \textbf{30.90} & \textbf{22.12} & \textbf{31.08} & \textbf{15.81} & \textbf{31.01} & \textbf{21.96} & \textbf{31.45} & \textbf{15.28} & \textbf{30.82} \\
\bottomrule
\end{tabular}}
\label{tab:sta_performance}
\vskip -0.1in
\end{table*}

\begin{figure*}[thbp!]
\vskip 0.1in
    \centering    
    \includegraphics[width=0.8\linewidth]{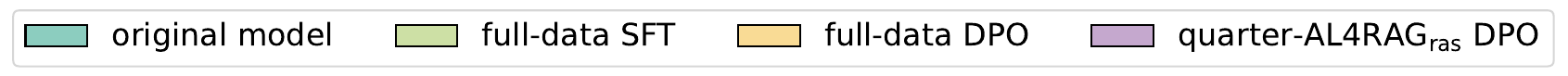}\\
    \begin{tabular}
    {@{\extracolsep{\fill}}c@{\extracolsep{\fill}}c@{\extracolsep{\fill}}c@{}}
        \includegraphics[width=0.32\linewidth]{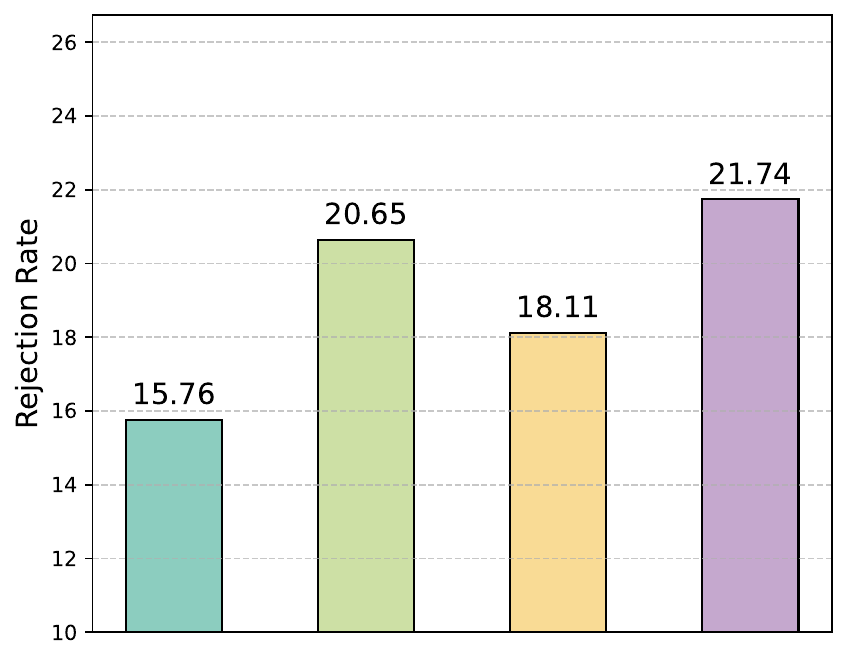} &
        \includegraphics[width=0.32\linewidth]{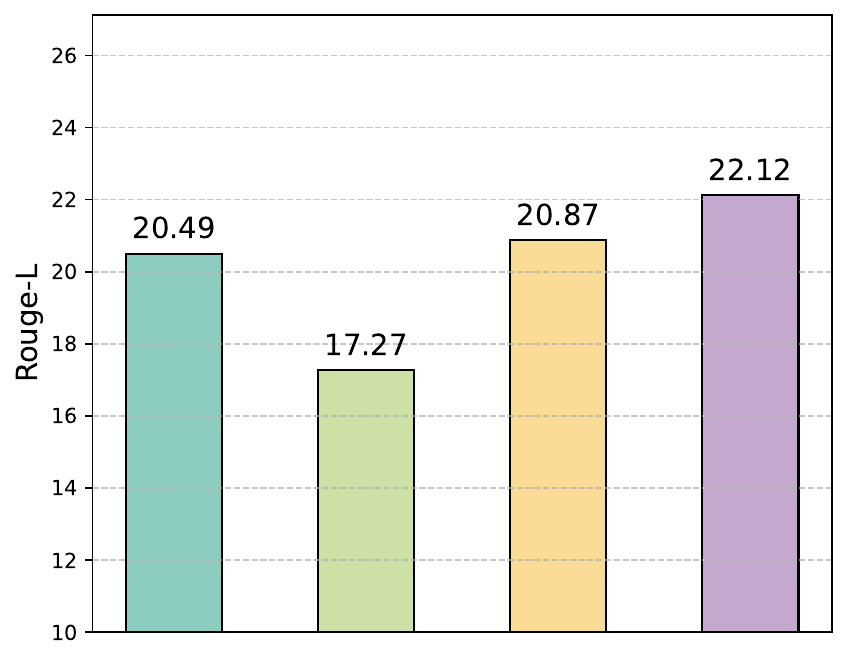} &
        \includegraphics[width=0.32\linewidth]{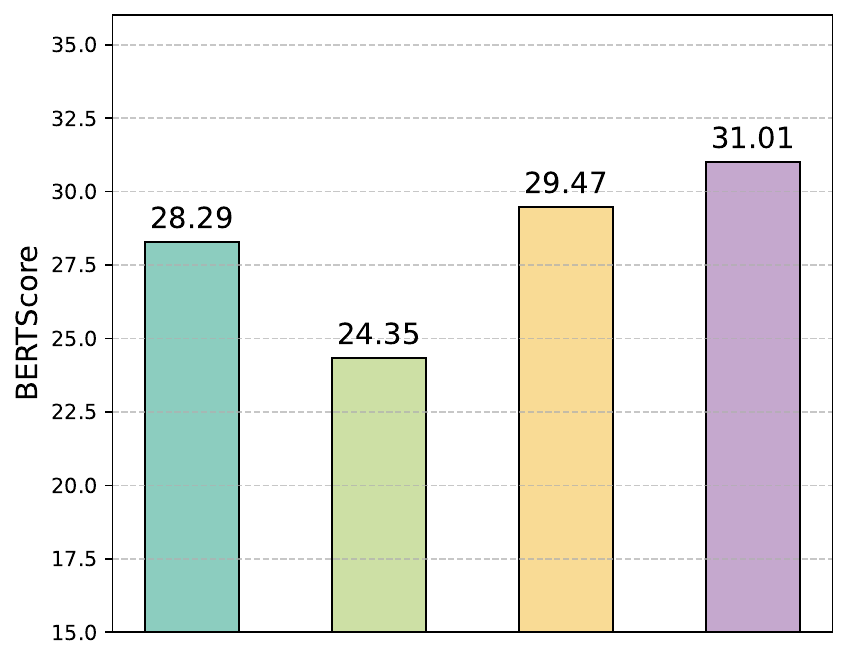}\\
        (a) Rejection Performance &
        (b) Stability Performance &
        (c) Stability Performance \\
    \end{tabular}
    \caption{The performance of the original model, the full-data SFT model, and the full-data DPO-trained model, with the performance of the model trained via DPO with 25\% data selected by our method as comparison. (a) Rejection rate of rejection performance; (b) Rouge-L of stability performance; (c) BERTScore of stability performance.}
    \label{fig:va_models}
\vskip -0.1in
\end{figure*}

\begin{table}[h]
\centering
\caption{Ablation study under a quarter of the data volume. The best results are \textbf{bolded}, and the second-best results are \underline{underlined}.}
\vskip 0.1in
\resizebox{0.8\columnwidth}{!}{
\begin{tabular}{@{}lccccccc@{}}
\toprule
\multirow{2}{*}{\textbf{Algorithms}} & \multicolumn{1}{c}{\textbf{Rejection}} & \multicolumn{2}{c}{\textbf{Stability}} \\
\cmidrule(lr){2-2} \cmidrule(lr){3-4}
 & \textbf{Rejection Rate} & \textbf{Rouge-L} & \textbf{BERTScore} \\
\midrule
AL4RAG\textsubscript{ras} & \textbf{21.74} & \textbf{22.12} & \textbf{31.01} \\
-q & 16.85 & \underline{21.20} & \underline{29.23} \\
+a & \underline{17.93} & 20.87 & 28.42 \\
\bottomrule
\end{tabular}}
\label{tab:ablation}
\vskip -0.1in
\end{table}

\begin{table}[h]
\centering
\caption{The impact of different vectorization methods on model performance under a quarter of the data volume. The best results are \textbf{bolded}, and the second-best results are \underline{underlined}.}
\vskip 0.1in
\resizebox{0.85\columnwidth}{!}{
\begin{tabular}{@{}lccccccc@{}}
\toprule
\multirow{2}{*}{\textbf{Methods}} & \multicolumn{1}{c}{\textbf{Rejection}} & \multicolumn{2}{c}{\textbf{Stability}} \\
\cmidrule(lr){2-2} \cmidrule(lr){3-4}
 & \textbf{Rejection Rate} & \textbf{Rouge-L} & \textbf{BERTScore} \\
\midrule
TF-IDF & \textbf{21.74} & \textbf{22.12} & \textbf{31.01} \\
Sentence-BERT & \underline{18.84} & \underline{21.83} & 29.97 \\
stella & 18.48 & 21.57 & \underline{30.23} \\
\bottomrule
\end{tabular}}
\label{tab:vec_method}
\vskip -0.1in
\end{table}

Table~\ref{tab:rej_performance} shows the refusal performance of different algorithms on hallucination-prone queries under data proportions of 12.5\%, 25\%, and 50\%.  
Our method consistently outperforms all baselines across all data ratios. Among IDDS-based variants, only ours maintains superior performance across all proportions. While Naive IDDS performs well at 12.5\%, its performance declines as data increases, being surpassed by more baselines. In contrast, our improvements enable robust performance by accurately measuring sample distances and identifying diverse samples.  

Interestingly, as data proportion increases, our method's performance slightly declines, while baselines improve. This suggests our method effectively selects the most impactful samples under smaller data proportions, reducing noise and redundancy. Figure \ref{fig:va_models}(a) compares rejection performance across models: the original model, a fine-tuned model (100\% data), a DPO-trained model (100\% data), and a DPO-trained model (25\% data, selected by our method). The model trained via DPO with data selected by our method achieves the best refusal performance, and the SFT model performs second-best, while the model trained via DPO with 100\% data performs worst after the original model. This highlights that fine-tuning equips models with the ability to reject answering hallucination-prone queries. Moreover, DPO training with a small amount of high-quality data further improves this capability, while training with a noisy dataset greatly weakens it. Compared to baselines, our method is the only one that consistently outperforms SFT in all data proportions, highlighting the superiority.

\subsubsection{Stability Results}

Table \ref{tab:sta_performance} presents the performance of various methods in improving the stability of correct answers, highlighting the effectiveness of different data selection strategies. 
Our method consistently outperforms all baselines across various data proportions. Notably, similar to the rejection task, it achieves peak performance at a data ratio of 12.5\%, a trend also observed in both IDDS-variant methods. This suggests that these methods can effectively identify informative samples even with a limited amount of data, ensuring efficient model optimization. Among them, our approach exhibits the most significant advantage, further demonstrating its effectiveness in balancing stability and refusal performance.  

Figures \ref{fig:va_models}(b) and \ref{fig:va_models}(c) compare the stability performance of different training approaches, including the original model, the SFT model trained on 100\% of the data, a DPO-trained model using 100\% of the data, and a DPO-trained model using only 25\% of the data selected by our method. While fine-tuning achieves strong rejection capabilities, it significantly reduces stability compared to the original model, indicating a trade-off between the two objectives. In contrast, DPO training with 100\% of the data reduces rejection performance compared to the SFT model but significantly improves stability, suggesting that optimizing for preference learning helps maintain correct answers. Furthermore, DPO training with 25\% data selected by our method improves stability without compromising rejection performance, demonstrating that our approach effectively balances both goals and enhances the model’s overall robustness.

\subsection{Ablation Study} \label{sec:ablation}

For the ablation experiment, we used a quarter of the dataset to compare performance under different conditions, with the results presented in Table \ref{tab:ablation}.  
Using only reference similarity ($ras-q$) as the overall sample similarity achieves stability performance comparable to our method but significantly underperforms in refusal accuracy. This is because the reference, as the most informative attribute, helps select diverse samples, enhancing stability. However, ignoring the influence of queries fails to distinguish samples with similar references but different questions. In such cases, the model may correctly answer some queries while failing others, making it difficult to determine when to refuse an answer, ultimately leading to weaker refusal performance.

Incorporating the answer ($ras+a$) slightly improves refusal performance compared to using only the reference but still falls short of our method, while its stability performance is the worst among the three approaches. Since responses are generated based on queries and references, most of their content already exists within these inputs, even if the response itself is incorrect. This redundancy hinders effective sample differentiation, leading to diverse but lower-quality selections, degrading the model’s performance in both tasks.

\subsection{Impact of different vectorization methods}

To explore the influence of different vectorization methods on the model performance, we selected two pre-trained text embedding models for comparison. One is \textit{Sentence-BERT} \cite{reimers2019sentence}, which is widely used in AL research. The other one is \textit{stella}\footnote{https://huggingface.co/dunzhang/stella\_en\_1.5B\_v5}, a 1.5B model ranked highly on the MTEB \cite{muennighoff2022mteb}  leaderboard\footnote{https://huggingface.co/spaces/mteb/leaderboard}. Table \ref{tab:vec_method} shows the performance of TF-IDF and these two text embedding models with a quarter of the data.

The two pre-trained models demonstrate comparable performance across tasks but slightly underperform compared to TF-IDF in both stability and rejection tasks. This discrepancy likely stems from fundamental differences in their approach: pre-trained models focus on capturing deep semantic relationships, while TF-IDF emphasizes surface-level text form and structure. This distinction allows TF-IDF to better differentiate samples with similar meanings but varying expressions and labels, making it particularly effective at identifying fine-grained variations. These nuanced samples are highly valuable for training, as they enhance the model’s ability to recognize subtle distinctions in responses. Consequently, TF-IDF's strength in capturing these textual differences contributes to its superior performance in both stability and rejection tasks, highlighting the importance of structural awareness in sample selection.

\section{Conclusion}

In this work, we introduce AL4RAG, the first AL framework for RAG, proposing an effective selection strategy tailored to the unique data patterns of RAG. To improve sample differentiation, we develop retrieval-augmented similarity (ras), enabling more accurate measurement of sample distances. Additionally, we expand the RAGTruth dataset and construct the first human preference dataset for RAG, allowing models to handle both hallucination-prone and answerable queries effectively. Extensive AL-driven optimization on the constructed dataset demonstrates that our approach consistently outperforms baselines, enhancing both response stability and the model’s ability to reject unreliable queries. These contributions provide a strong foundation for advancing RAG-based learning and optimizing LLMs with external knowledge.

\section*{Impact Statement}

There may be potential social consequences of our work. In the context of data privacy, the use of historical conversation records is a critical concern. As we rely on these records to train and optimize the model, there is a potential risk of exposing sensitive user information. If proper security measures are not in place, this could lead to privacy breaches, where personal details, opinions, or interactions captured in the conversations might be accessed without authorization.

\nocite{DBLP:conf/iclr/WangXLX0024,DBLP:conf/kdd/WangZH00024,DBLP:conf/www/WangJZHR00024,DBLP:journals/corr/abs-2412-13437,DBLP:journals/corr/abs-2412-13840}

\bibliography{example_paper}
\bibliographystyle{icml2025}



\end{document}